\documentclass[11pt]{article}

\pdfoutput=1

\usepackage{graphicx}
\usepackage[algoruled,vlined,linesnumbered]{algorithm2e}
\usepackage{amsmath}
\usepackage{amssymb}

\newtheorem{df}{Definition}

\newcommand{\bt}{\begin{theorem}\em}
\newcommand{\et}{\end{theorem}}

\newcommand{\nin}{\noindent}

\newcommand{\bea}{\begin{eqnarray}}
\newcommand{\eea}{\end{eqnarray}}

\newcommand{\bdf}{\begin{df}\em}
\newcommand{\edf}{\end{df}}
 
\newcommand{\ben}{\begin{enumerate}}
\newcommand{\een}{\end{enumerate}}

\newcommand{\mpi}{\widetilde{\pi}}
\newcommand{\ab}{\bar{a}}
\newcommand{\co}{\bar{c}}

\newcommand\T{\rule{0pt}{2.6ex}}       
\newcommand\B{\rule[-1.2ex]{0pt}{0pt}} 

\DeclareMathOperator*{\argmax}{arg\,max}

\begin{document}

\title{Flow for Meta Control}

\author{%
Vadim Bulitko \\
\small Department of Computing Science, University of Alberta\\
\small Edmonton, Alberta, T6G 2E8, CANADA \\
\small {\tt bulitko@ualberta.ca}}

\maketitle

\begin{abstract}

The psychological state of flow has been linked to optimizing human performance. A key condition of flow emergence is a match between the human abilities and complexity of the task. We propose a simple computational model of flow for Artificial Intelligence (AI) agents. The model factors the standard agent-environment state into a self-reflective set of the agent's abilities and a socially learned set of the environmental complexity. Maximizing the flow serves as a meta control for the agent. We show how to apply the meta-control policy to a broad class of AI control policies and illustrate our approach with a specific implementation. Results in a synthetic testbed are promising and open interesting directions for future work.

\smallskip \nin {\bf Keywords:} flow, computational model, meta control.
\end{abstract}

\section{Introduction}

In psychology the state of flow has been linked to optimizing human problem-solving performance~\cite{flow2008}. The key condition of flow emergence is a match between the agent's abilities and  complexity of the problem the agent is solving. In people, experiencing flow manifests itself as a feeling of happiness which attracts people to tasks that fully engage their abilities. Thus maximizing flow can be viewed as a guide to improving performance. 

In this paper we present a simple computational approach of equipping Artificial Intelligence (AI) agents with a sense of flow. To do so we factor the  state in the usual agent-in-an-environment framework~\cite{russellNorvig} into a self-reflective part (the agent's abilities) and an objective part (environmental complexity).  The AI agent's control policy is then augmented with a flow-maximizing meta-control module which guides the agent to the areas of the environment where the agent's abilities match the environmental complexity. There the agent's base control policy has a potential to perform well.

In the past this approach was applied to Reinforcement Learning (RL) agents~\cite{RL} where the flow maximization was implemented via, essentially, an additional flow reward signal~\cite{cogsys2012}. The agent then maximized a linear combination of its usual cumulative reward and the expected value of the flow return. The flow return was defined as a reciprocal of the absolute difference between the agent's ability (a scalar) and the environmental complexity (also a scalar). Both scalar variables were hand-coded into the agent. In a simple synthetic environment, a flow-maximizing RL agent outperformed a baseline~\cite{cogsys2012}. 

In this paper we address the two primary limitations of the published work: (i) the assumption of an RL agent architecture and (ii) accurate environmental complexity being hand-coded into the agent. Thus,  we  apply the flow-maximizing meta control to a broad class of base control policies, extending the applicability beyond RL. Second, we propose a way for the agent to learn the environmental complexity by observing other agents. We illustrate our ideas on a simple synthetic problem and discuss its possible extensions.

\section{Problem Formulation}\label{sec:formulation}

 \subsection{Restrictions on the Problem}

To apply the flow-driven meta control to AI agent architectures beyond RL we will impose certain restrictions on the environment the  agents operate in. We represent the environment as a Markov Decision Process (MDP) which consists of a set of {\em states} $S$, a set of {\em actions} $A$ and a {\em transition probability} function $p$. We assume a partitioning of the state set $S$ into $n+1$ subsets, or {\em levels}, $L_i$:
\bea 
S &=& \bigcup_{i=0}^n L_i, \\ 
\forall i,j & & \left[ i, j \in \{0,\dots,n\} \And i \neq j \to L_i \cap L_j = \emptyset \right].
\eea
The agent's start state $s_0$ is at level $0$ ($s_0 \in L_0$). At each discrete time step $t$ the agent takes an action $\pi(s_t) \in A$ which  brings the agent to the next state $s_{t+1}$. Formally, the state $s_{t+1}$ is drawn from $S$ according to the transition probability $p(s_{t+1} | s_t,\pi(s_t))$. We denote this as: $s_{t+1} \xleftarrow{\pi(s_t),p} s_t$.

While this formulation allows for episodic as well as non-episodic tasks, in the rest of the paper we work with a special case of this problem: an episodic stochastic shortest path. Specifically, the agent's task is to reach the highest level $L_n$ quickly and reliably. The agent starts in the start state $s_0 \in L_0$ and runs until either reaching a state at level $L_n$ or dying (i.e., transitioning to a designated state $s^\dagger$). We incorporate the death state into the MDP as follows:
\bea 
S &=& \left\{s^\dagger\right\} \cup \bigcup_{i=0}^n L_i; \ \ \ \ \ \ \ \ \ 
s^\dagger \notin \bigcup_{i=0}^n L_i.
\eea 

\subsection{Performance Measure}

To quantify the agent's performance, we will reward the agent with its current level $i$ at each state $s_t \in L_i$. If the agent dies before it reaches $L_n$ then it forfeits its entire accumulated reward. Thus the agent's life-time {\em return} is:
\bea
R_{T_{\max}} = 
\begin{cases}
\sum_{t=0}^{T_{\max}} i, & \text{if the agent has reached $L_n$}, \\
0, & \text{if the agent has died prior to $L_n$ (i.e., reached $s^\dagger$)}.
\end{cases} 
\eea

\begin{figure}[htbp]
\begin{center}
\includegraphics[width=7cm]{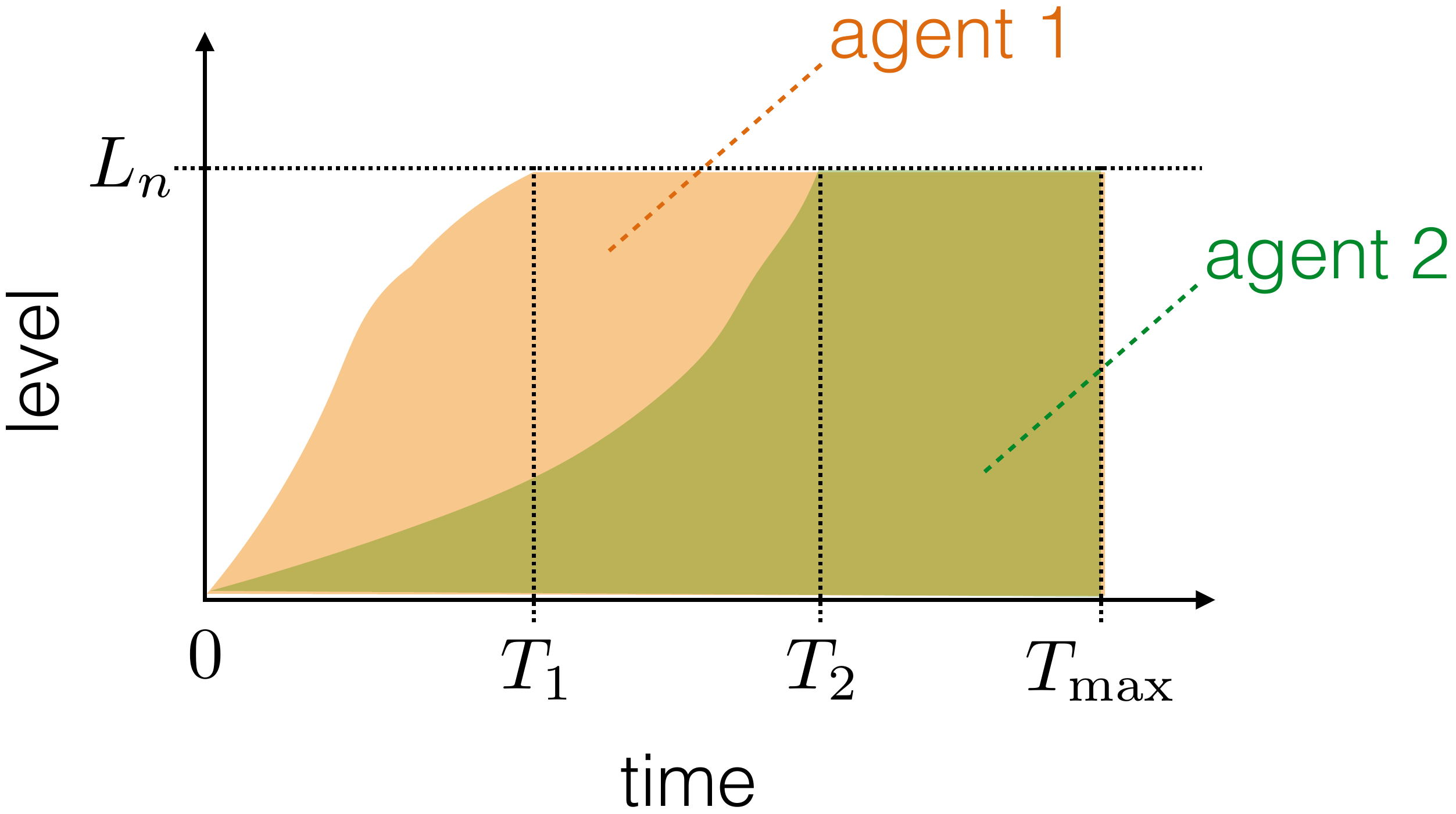}  
\caption{Comparing performance of two agents.}
\label{fig:performanceComparison}
\end{center}
\end{figure}

Suppose an agent reached $L_n$ at time $T$. If $T_{\max}$ happens to be above $T$ then we continue to reward the agent with $n$ for each time step between $T+1$ and ${T_{\max}}$. This allows us to compare two agents as shown in Figure~\ref{fig:performanceComparison}. There, the first agent reaches $L_n$ at time $T_1$ and then remains at that level until the time $T_{\max}$, collecting the reward of $n$ at each time step between $T_1$ and $T_{\max}$. The second agent reaches $L_n$ at time $T_2$ and also receives the reward of $n$ for each $t \in \{T_2+1, \dots,T_{\max}\}$. The returns the two agents collect are the areas under the level-ascension curves. 

Note that while we use rewards to define the agent's performance we do not assume that the agent has access to the rewards. Thus, we are not restricting the agent architecture to Reinforcement Learning as was done in the past~\cite{cogsys2012}.

\subsection{Restrictions on the Agent Design}

We consider the problem of meta control by assuming that the agent already has a {\em control policy} $\pi : S \to A$. We restrict it so that it never moves the agent between levels: $\forall 0 \le i,j \le n \forall s \in L_i \forall s' \in L_j \left[ i \neq j \to p(s'|\pi(s),s) = 0 \right]$. 
Moving between levels is accomplished with a {\em meta-control policy} $\mpi : S \to A$ whose actions either keep the agent in the same state or move it to a state in another level or cause its death: $
\forall 0 \le i \le n \forall s,s' \in L_i \left[ s \neq s' \to p(s'|\mpi(s),s) = 0 \right]$.

Any agent can use both the control policy $\pi$ and the meta-control policy $\mpi$ as shown in Algorithm~\ref{alg:overallOperation}. In the main loop (line~\ref{al:loop} of the algorithm) a sequence $(s_0,s_2,\dots,s_T)$ of states ending in either the agent's death or in the target level $L_n$ is generated by successively applying the meta control $\mpi$ (line~\ref{al:inter}) and the control $\pi$ (line~\ref{al:intra}). 

\begin{algorithm}[htbp]
\DontPrintSemicolon
\label{alg:overallOperation}
\caption{Agent Operation}

\SetKwInOut{Input}{input}
\SetKwInOut{Output}{output}
\Input{MDP $(S,A,p)$, start state $s_0$, control policy $\pi$, meta-control policy $\mpi$}
\Output{trajectory $(s_0,s_2,\dots,s_T), s_T \in \{s^\dagger\} \cup L_n$}

$t \gets 0$ \;

\While{$s_t \notin \{s^\dagger\} \cup L_n$}{\label{al:loop}
$s' \xleftarrow{\mpi(s_t),p} s_t$ \;
\label{al:inter}
$s_{t+1} \xleftarrow{\pi(s'),p} s'$ \;
\label{al:intra}
$t \gets t+1$\; 
}
\end{algorithm}

\section{Related Work}\label{sec:relatedWork}

Meta-control policies have been an important element of AI since its early days. The classic A* algorithm uses a heuristic to control its search at the base level and breaks ties towards higher $g$-costs at the meta-control level. Pathfinding algorithms often use heuristic search (e.g., A*) as the base control policy but meta-control it with another search~\cite{SturtevantAIIDE07,Bulitko:07-jair} or case-based reasoning~\cite{knn}. Hierarchical control can also be used to solve MDPs more efficiently~\cite{prlrtsMDP}.

Existing meta-control policies are diverse and specific to the underlying control policy. Thus they cannot always be ported across different base control policies/architectures. We address this shortcoming in the following by suggesting a single simple meta-control policy that explicitly factors the agent's state into a self-reflective part (the agent's abilities) and the objective, societally learned, part (the environmental complexity). Doing so de-couples our meta-control approach from the underlying control policy and thus makes it applicable to a broad range of AI architectures.


\section{Our Approach}


\subsection{Intuition}

As argued in the introduction, flow-maximizing agents attempt to position themselves in the areas of the environment where their abilities match the complexity of the environment. In our formalization, the areas of the environment are levels and the positioning happens via a meta-control policy which guides the agent to the appropriate level. Hence  reasoning about flow happens within  meta control. 

Generally speaking, giving an agent the ability to position itself in the area of the environment of its own choice may interfere with the agent's reaching a designer-specified goal. In this study we make an assumption that the environment is such that building up the agent's abilities at lower levels makes the agent more capable of tackling higher levels, all the way to the goal level $L_n$. This assumption holds for many common tasks (e.g., sports). 

In line with previous work on flow in AI~\cite{cogsys2012}, we define the degree of flow as the quality of the match between the agent's abilities and the environmental complexity. Then the flow-maximizing meta-control policy guides the agent to the level of the environment for which the agent is currently most suited. There, the control policy has the best chance to maximize its performance. As the agent's abilities increase over time, flow-maximizing meta-control guides the agent to higher levels of the environment.

The complexity of a level can be determined via social learning: the agent observes performance of other agents which have visited the level before it. The minimum abilities that were sufficient to reach the highest level starting at the given level are then taken as the complexity of that level.

\subsection{Algorithmic Details}
\label{sec:algDetails}


\subsubsection{Agent's Abilities.}

\begin{figure}[htbp]
\begin{center}
\includegraphics[width=\textwidth]{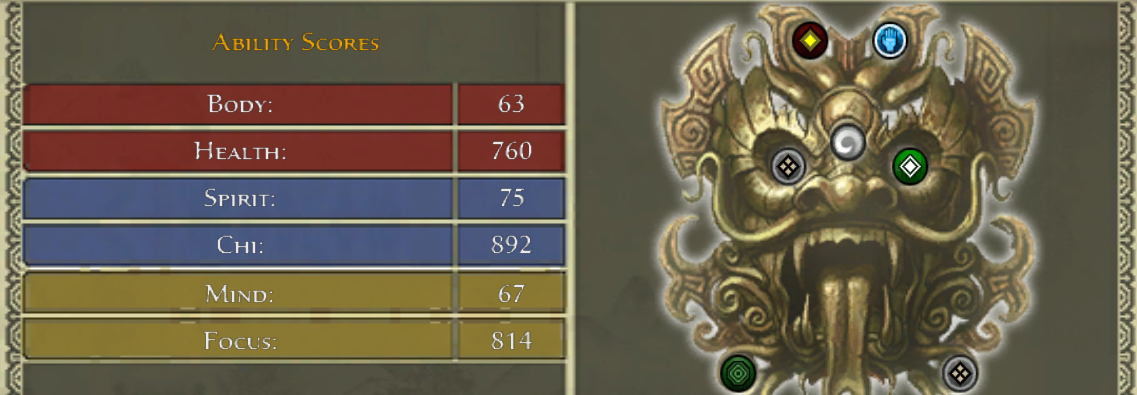}  
\caption{Agent's abilities in the game of {\em Jade Empire}.}
\label{fig:jadeEmpire}
\end{center}
\end{figure}

We define the agent's abilities at time $t$ as a $k$-component real-valued vector $\ab_t \in \mathbb{R}^k$. For instance, the player in the video game {\em Jade Empire} can have its body and mind abilities quantified at $63$ and $67$ (Figure~\ref{fig:jadeEmpire}). In our notation this would represented as $\ab = (63,67)$. 

\subsubsection{Problem Complexity.}

The problem complexity $\co(L_i)$ at the level $L_i$ is defined as the minimum agent's abilities needed to solve the problem (i.e., reach the final level $L_n$) from the level $L_i$ with a high probability.

The agent can estimate the complexity of level $L_i$ by observing other agents at that level and recording their abilities. It then filters out all such agents that did not reach $L_n$ and selects the minimum among the remaining abilities. To illustrate: suppose three agents operated at level $L_i$. Their abilities were $(63,67)$, $(40,70)$ and $(10,35)$. Suppose the last agent died before reaching $L_n$ while the first two survived. The complexity of level $L_i$ is then estimated as per-component minimum of the vectors $(63,67)$ and $(40,70)$ which is $(40, 67)$.

This approach is based on three assumptions. First, we assumed that higher values in the ability vector indicate a higher probability of reaching $L_n$. Second, we assumed that the complexity $\co$ of a level is uni-modal and thus has a single vector expressing the required abilities. Third, we assumed that the collected set of other agents' abilities is sufficient to cover the space of abilities well enough to reliably estimate the required abilities for a level. We will challenge some of these assumptions in the future work section.

\subsubsection{Degree of Flow.}

The model of flow we use is an extension of existing work~\cite{cogsys2012}. The {\em degree of flow} is defined as the divergence between the agent's abilities and the complexity of the level the agent is at. Mathematically, the degree of flow is $F(\ab,\co) = 1 / \left(|\ab - \co| + \xi \right)$ where $|\ |$ is the Euclidean distance and $\xi > 0$ is a constant to avoid division by zero. $F$ reaches its maximum value of $1/\xi$ when the agent's abilities match the level complexity precisely. 
To illustrate: an agent with the ability vector $(63,67)$ operating at level with the complexity $(40,67)$ will experience flow to the degree $F = 1 / (\sqrt{(63-40)^2 + (67-67)^2}+\xi) \approx 0.04$.

\subsubsection{Meta-control Policy.}
\label{sec:metacontrol}

If the agent is in the state $s_t$, its meta-control policy $\mpi$ considers its current level $L_t \ni s_t$ as well as all neighboring levels in the interval $[L_t - \Delta_t, L_t + \Delta_t]$ where $\Delta_t$ is the radius of the neighbourhood. The policy selects the target level as:
\bea
L_{t+1} =  \argmax_{L \in [L_t - \Delta_t, L_t + \Delta_t]} F(\ab_t,\co(L))
\label{eq:metaPolicy}
\eea
Once the next level $L_{t+1}$ is computed, the meta-control policy $\mpi$ outputs the action the moves the agent to that level (line~\ref{al:inter} in Algorithm~\ref{alg:overallOperation}).


\section{Illustration}

We illustrate our approach by implementing it in a simple testbed. Its synthetic nature gives us a fine control over the environment enabling a clear presentation.

\subsection{The Testbed}

We consider the agent's abilities and the environmental complexity to be scalars (i.e., $k=1$). Further, we assume that the agent's ability is simply its age: $\forall t \left[ \ab_t = t \right]$. We focus exclusively on meta control by having each state to be its own level: $\forall L \exists s \left[ L = \{ s \} \right]$ with $\mpi$ being the only control the agent uses. The probability of dying is defined as:
\bea
p(s^\dagger | s_t, \mpi(s_t)) = 
\begin{cases}
p^\dagger, & \text{if $\ab_t \ge \co(L_t)$}, \\
\min\left\{1,p^\dagger + \tanh(\co(L_t) - \ab_t)\right\}, & \text{otherwise},
\end{cases}\label{eq:prDeath1} 
\eea
where $p^\dagger \in [0,1]$ is the probability that any agent dies at any given time step, regardless of its abilities and the problem complexity. If the agent is able enough for it level $L_t$ (i.e., $\ab_t \ge \co(L_t)$) then $p^\dagger$ is the sole contributor to the probability of dying. If the agent's abilities are below the level's complexity (i.e., $\ab_t < \co(L_t)$) then the probability of dying is the sum of the ambient death probability $p^\dagger$ and the probability of dying from a lack of the abilities: $\tanh(\co(L_t) - \ab_t)$. The sum is capped at $1$ by taking the minimum. 


If the agent does not die at a given time step then the meta control $\mpi$ is able to reliably bring it to any level it specifies. 

\subsection{Experiments}

For the baseline non-flow agents we set the meta-control policy to bring the agent to level $\alpha \cdot t$ where $\alpha$ as a parameter. For instance, a baseline agent with $\alpha = 0.5$ will be at level $5$ at time step $t = 10$ (if it survives that long).

\begin{figure}[htbp]
\begin{center}
\includegraphics[height=5.35cm]{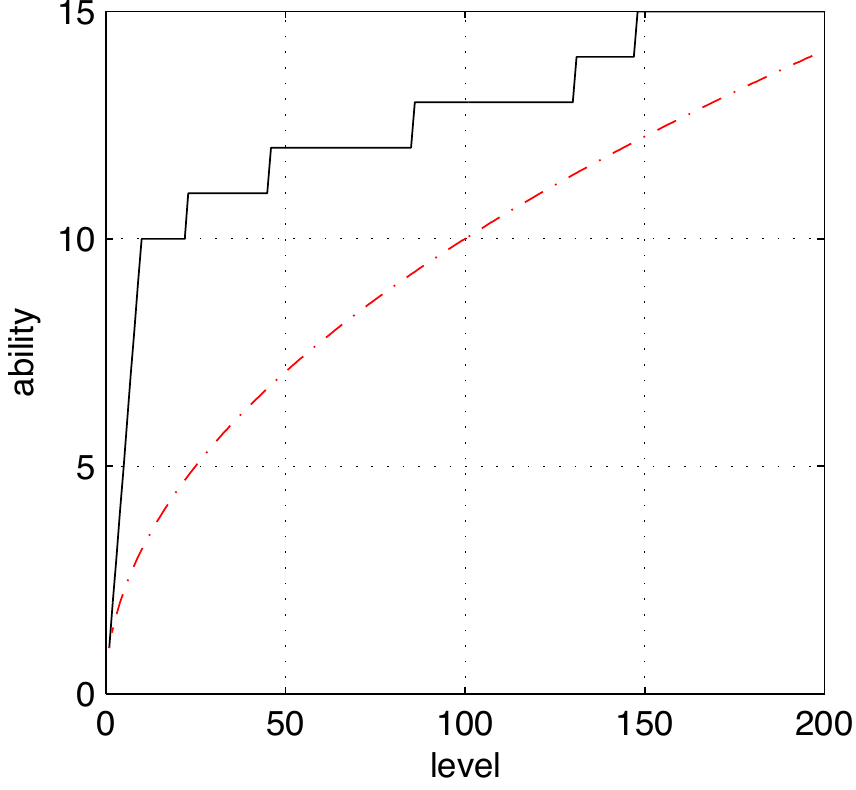}  
\hspace{0.1cm}
\includegraphics[height=5.35cm]{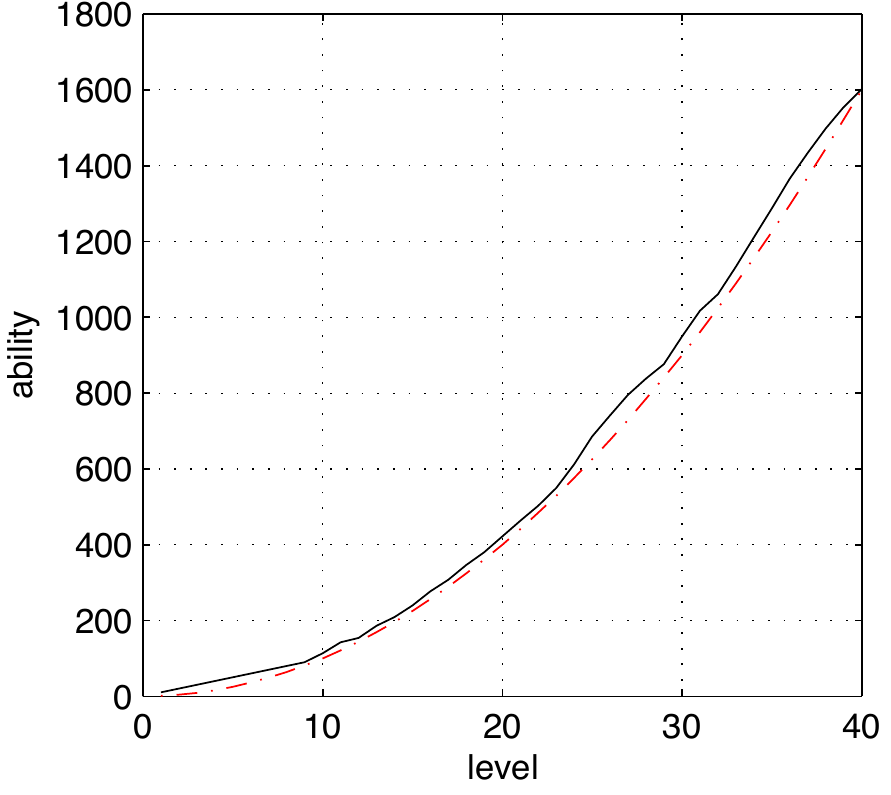}  
\caption{Actual (dashed line) and data-mined (solid line) complexities for $\co(L) = \sqrt{L}$ (left) and $\co(L) = L^2$  (right). For the square root complexity the ambient death probability is $p^\dagger = 0.001$. For the quadratic complexity $p^\dagger = 0.0001$. In both cases we used $10^4$ probes and removed $\rho = 0.1\%$ of lowest ability points.}
\label{fig:minedComplexity}
\end{center}
\end{figure}

As described earlier in the paper, the flow agents first learn the complexity of each level by observing multiple {\em probe agents}. The probe agents behave as a baseline agent parameterized by $\alpha$ until a  randomly selected level $l'$. Then they choose the $\alpha$ randomly and follow it until a randomly selected level $l''$. At that point they once again randomize their $\alpha$. Effectively these agents are piece-wise linear with two joint points. The complexity at level $L$ is then defined as the lowest ability observed at level $L$ among any probe agent which went on to reach level $L_n$. We remove the lowest $\rho$ percent of the ability data per level as outliers (i.e., the probe agents that did not have the abilities necessary to reach $L_n$ but reached it nevertheless by luck). Figure~\ref{fig:minedComplexity} compares  data-mined and  actual complexities of two different environments: $\co(L) = L^2$ and $\co(L) = \sqrt{L}$.

Once the complexity is approximated via taking the per-level minima of the recorded probe agent abilities, the flow agents use the flow-maximizing meta policy to advance them through the levels. In effect,  the flow-maximizing agents attempt to follow the mined complexity curve.\footnote{To simplify the illustration we made the levels continuous. Our meta-control policy tried advancing the agent's current level in small increments ($0.001$) until it found the maximum of the flow function.} The results are found in Table~\ref{tab:results}. 

\begin{table}[htdp]
\caption{Flow-maximizing versus baseline agents: mean return $\pm$ standard error.}\label{tab:results}
\begin{center}
\begin{tabular}{c|c|c}\hline
{\bf Environment} & {\bf Baseline} & {\bf Flow maximizing} \T\B \\  \hline
$\co(L) = \sqrt{L}$, $L \in [0,200]$ & $6.38 \times 10^5\pm 9554$ & $7.86 \times 10^5 \pm 3068$ \T\B \\
$\co(L) = L^2$, $L \in [0,40]$ & $1.06 \times 10^5\pm 1481$ & $1.18 \times 10^5 \pm 1518$ \T\B \\ \hline
\end{tabular}
\end{center}
\end{table}

\begin{figure}[htbp]
\begin{center}
\includegraphics[height=5.7cm]{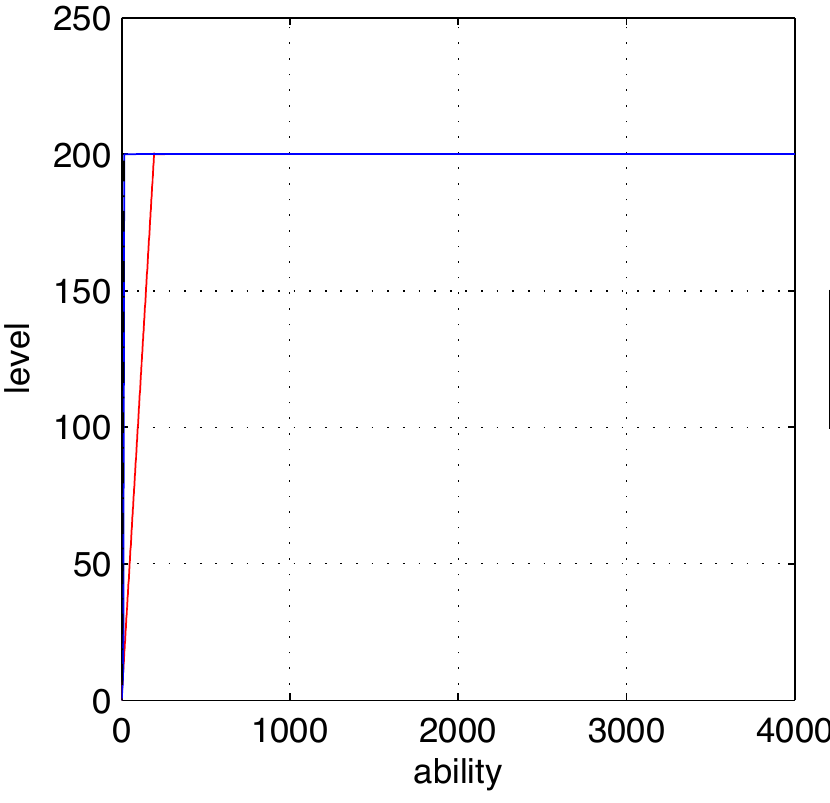}  
\hspace{0.1cm}
\includegraphics[height=5.7cm]{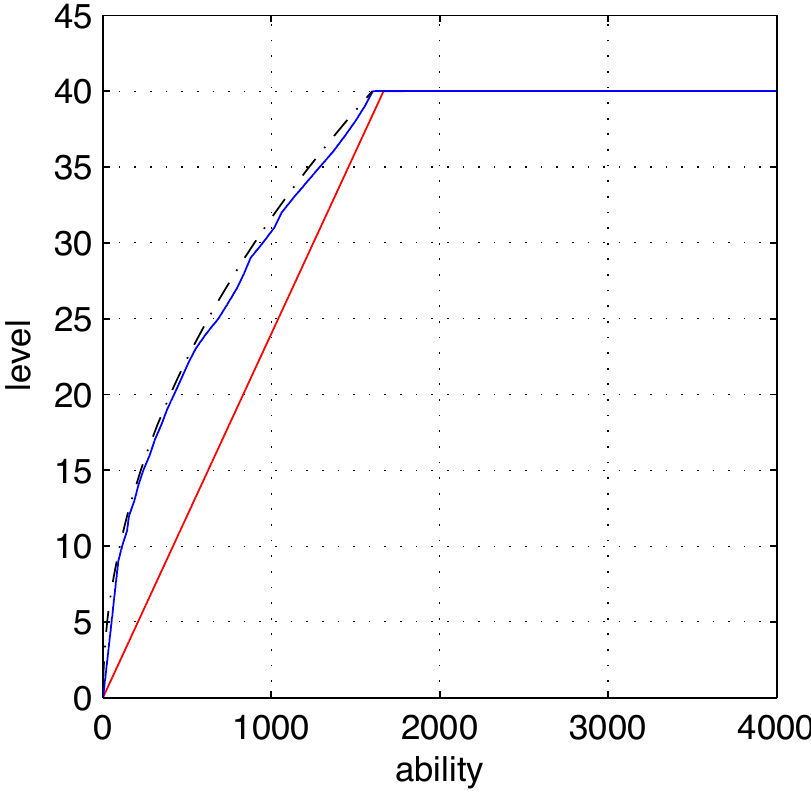}  
\caption{A single trial of baseline and flow-maximizing agents for the square root complexity (left) and the quadratic complexity (right) environments. The dashed line is the actual complexity. The straight line is the best baseline agent while the other solid line is the flow agent.}
\label{fig:results}
\end{center}
\end{figure}

For the square root complexity curve $\co(L) = \sqrt{L}$, the baseline agents were tried with $10$  values of the $\alpha$ parameter, tabulated from $[0.5, 3]$. For each $\alpha$ value we ran $1000$ trials. The flow agents used the data-mined complexity curve from Figure~\ref{fig:minedComplexity} and we also ran $1000$ trials. The returns were computed until $T_{\max} = 4001$. As per Table~\ref{tab:results} the baseline agents achieved the average return of $6.38 \times 10^5\pm 9554$ for the best value of $\alpha = 1.03$. The flow agents outperformed them with the average return of $7.86 \times 10^5 \pm 3068$. A single trial of the baseline agent with that $\alpha$ value is shown in Figure~\ref{fig:results} (left), together with a single trial of the flow agent. 

For the quadratic complexity curve $\co(L) = L^2$ the baseline agents were tried with $10$  values of the $\alpha$ parameter, tabulated from $[0.01, 0.028]$. For each $\alpha$ value we ran $1000$ trials. The flow agents used the data-mined complexity curve from Figure~\ref{fig:minedComplexity} and we also ran $1000$ trials. The returns were computed until $T_{\max} = 4001$. As per Table~\ref{tab:results} the baseline agents achieved the average return of $1.06 \times 10^5\pm 1481$ for the best value of $\alpha = 0.024$. The flow agents outperformed them again with the average return of $1.18 \times 10^5 \pm 1518$. A single trial of the baseline agent with that $\alpha$ value is shown in Figure~\ref{fig:results} (right), together with a single trial of the flow agent.

\section{Future Work}

By selecting states where the agent's abilities match the environmental complexity the flow-maximizing agents outperformed the baseline agents. We used this predictable result to illustrate the approach and offer a number of interesting future research directions. 

In defining the agent's ability vector, it will be interesting to try automated feature selection methods to identify relevant features. In defining the level complexity one can attempt to use automated clustering methods to deal with multi-modality. For instance, many video games allow different character classes (e.g., strong and slow versus weak and fast) to be equally successful. By clustering the observed data first and then taking the minimum in each cluster the agent will compute several required ability vectors per level. If data from previous agents are unavailable then the agent can attempt to estimate the level complexity from its own performance at the level. A particular promising direction may be the dynamics of the temporal-difference (TD) error~\cite{CP}. Alternatively, level complexity can be innate within agents, evolved over  generations thereby making flow-maximizing meta control an evolutionary adaptation. Finally, it will be interesting to see how well this model of flow correlates with human flow data.

\section{Conclusions}\label{sec:conclusions}

We proposed a simple psychology-inspired meta-control approach based on matching the agent's abilities and the environmental complexity. The approach is applicable to a broad spectrum of existing AI control policies. We factored the usual AI agent-environment state into a self-reflective and objective parts and applied social learning to determine the latter. We illustrated the approach with a specific implementation in a  simple synthetic testbed.

\section{Acknowledgments}

We are grateful to Th\'{o}rey Mar\'{i}usd\'{o}ttir and Matthew Brown for fruitful discussions. We appreciate funding from NSERC.

\bibliography{bibliography}
\bibliographystyle{splncs03}

\end{document}